\documentclass{ifacconf}

\usepackage{epsfig}
\usepackage{amsmath}
\usepackage{amssymb} 
\usepackage{longtable}
\usepackage{rotating}
\usepackage{multirow}
\usepackage{array}
\usepackage{graphicx}
\usepackage{mathrsfs}
\usepackage{verbatim}
\usepackage{color}
\usepackage{epstopdf}
\usepackage{url}
\usepackage{graphicx}
\usepackage{mathrsfs}
\usepackage{verbatim}
\usepackage{color}
\usepackage[noadjust]{cite}
\usepackage{epstopdf}
\usepackage{url}
\usepackage{natbib}        
\usepackage{amsmath} 
\usepackage{amssymb}  
\usepackage{bbm}
\usepackage{color}
\usepackage{graphics}
\usepackage{graphicx}
\usepackage{epsfig}
\usepackage{epstopdf}
\usepackage{amsfonts}
\usepackage{mathtools}
\usepackage{bm}
\usepackage{fancyhdr}
\usepackage{framed}
\usepackage{float}
\usepackage{cite}

\usepackage{tikz}
\usetikzlibrary{calc,positioning,shapes,shadows,arrows,fit}

\newcommand{\bs}{\boldsymbol}
\usepackage{pifont}

\newtheorem{theorem}{\textbf{Theorem}}

\newtheorem{problem}{Problem}

\newtheorem{proposition}{\textbf{Proposition}}
\newtheorem{remark}{Remark}
\newtheorem{definition}{\textbf{Definition}}

\begin{document}
	\begin{frontmatter}
		
		\title{A Robust Force Control Approach for Underwater Vehicle Manipulator
		Systems\thanksref{footnoteinfo}} 
		
		\thanks[footnoteinfo]{This work was supported by the ”ROBOCADEMY”, Marie Curie ITN Grant Agreement no FP7-608096 funded by the EU action ”$7^{th}$ Framework Programme - The 2013 People Work Programme - EC Call Identifier FP7-PEOPLE-	2013-ITN, Implementation Mode: Multi-ITN”}
		
		\author[First]{Shahab Heshmati-alamdari} 
		\author[Second]{Alexandros Nikou} 
		\author[First]{Kostas J. Kyriakopoulos} 
		\author[Second]{Dimos V. Dimarogonas}
		
		\address[First]{Control Systems Lab, Department of Mechanical Engineering, National Technical University of Athens, 9 Heroon Polytechniou Street, Zografou 15780. \\ E-mail: \{shahab, kkyria\}@mail.ntua.gr}
		
		\address[Second]{ACCESS Linnaeus Center, School of Electrical Engineering and KTH Center for Autonomous Systems, KTH Royal Institute of Technology, SE-100 44, Stockholm, Sweden. \\ E-mail: \{anikou, dimos\}@kth.se}
		
		\begin{abstract}                
In various interaction tasks using Underwater Vehicle Manipulator Systems (UVMSs) (e.g. sampling of the sea organisms, underwater welding), important factors such as: i) uncertainties and complexity of UVMS dynamic model ii) external disturbances (e.g. sea currents and waves) iii) imperfection and noises of measuring sensors iv) steady state performance as well as v) inferior overshoot of interaction force error, should be addressed during the force control design. Motivated by the above factors, this paper presents a  model-free control protocol for force controlling of an Underwater Vehicle Manipulator System which is in contact with a compliant environment, without incorporating any knowledge of the UVMS's dynamic model, exogenous disturbances and sensor's noise model. Moreover, the transient and steady state response as well as reduction of overshooting force error are solely determined by certain designer-specified performance functions and are fully decoupled by the UVMS's dynamic model, the control gain selection, as well as the initial conditions. Finally, a simulation study clarifies the proposed method and verifies its efficiency.
		\end{abstract}
		
		
	\end{frontmatter}
\section{Introduction}

In view of the development of autonomous underwater vehicles, the capability of such vehicles to interact with the environment by the use of a robot manipulator, had gained attention in the literature. Most of the underwater manipulation tasks, such as maintenance of ships, underwater pipeline or weld inspection, surveying, oil and gas searching, cable burial and mating of underwater connector, require the manipulator mounted on the vehicle to be in contact with the underwater object or environment. The aforementioned systems are complex and they are characterized by several strong constraints, namely the complexity in the mathematical model and the difficulty to control the vehicle. These constraints should be taken into consideration when designing a force control scheme. In order to increase the adaptability of UVMS, force control must be included into the control system of the UVMS. Although many force control schemes have been developed for earth-fixed manipulators and space robots, these control schemes cannot be used directly on UVMS because of the unstructured nature of the underwater environment.

From the control perspective, achieving these type of tasks requires specific approaches\citep{Siciliano_Sciavicco}. However, speaking about underwater robotics, only few publications deal with the interaction control using UVMS. One of the first underwater robotic setups for interaction with the environment was presented in \citep{Casalino20013220}. Hybrid position/force control schemes for UVMS were developed and tested in \citep{lane1,lane2}. However, dynamic coupling between the manipulator and the underwater vehicle was not considered in the system model. In order to compensate the contact force, the authors in \citep{kajita} proposed a method that utilizes the restoring force generated by the thrusters. In the same context, position/force \citep{lapierre}, impedance control \citep{cui1,cui2,cui3}  and  external force control schemes \citep{antonelli_tro, antonelli_cdc,Antonelli2002251} can be found in the literature. 

Over the last years, the interaction control of UVMS is gaining significant attention again. Several control issues for an UVMS in view of intervention tasks has been presented in \citep{Marani2010175}. In \citep{Cataldi2015524} based on the interaction schemes presented in \citep{antonelli_tro} and \citep{Antonelli2002251}, the authors proposed a control protocol for turning valve scenarios. Recent study \citep{moosavian} proposed a multiple impedance control scheme for a dual manipulator mounted on AUV. Moreover, the two recent European projects  TRIDENT(see, e.g. \citep{Fernández2013121},\citep{Prats201219},\citep{Simetti2014364}) and PANDORA (see, e.g. \citep{Carrera2014}, \citep{Carrera2015}) have given boost to underwater interaction with relevant results. 

In real applications, the UVMS needs to interact with the environment via its end-effector in order to achieve a desired task. During the manipulation process the following issues occur: the environment is potentially unknown, the system is in the presence of unknown (but bounded) external disturbances (sea currents and sea waves) and the sensor measurements are not always accurate (we have noise in the measurements). These issues can cause unpredicted instabilities to the system and need to be tackled during the control design. From the control design perspective, the UVMS dynamical model is highly nonlinear, complicated and has significant uncertainties. Owing to the aforementioned issues, underwater manipulation becomes a challenging task in order to achieve low overshoot, transient and steady state performance.

 Motivated by the above, in this work we propose a force - position control scheme which does not require any knowledge of the UVMS dynamic parameters, environment model as well as the disturbances. More specifically, it tackles all the aforementioned issues and guarantees a predefined behavior of the system in terms of desired overshoot and prescribed transient/steady state performance. Moreover, noise measurements, UVMS model uncertainties (a challenging issue in underwater robotics) and external disturbance are considered during control design. In addition, the complexity of the proposed control law is significantly low. It is actually a static scheme involving only a few calculations to
output the control signal, which enables its implementation on most of current UVMS. 
The rest of this paper is organized as follows: in Section 2 the mathematical model of UVMS and preliminary background are given. Section 3 provides the problem statement that we aim to solve in this paper. The control methodology is presented in Section 4. Section 5 validates our approach via a simulation study. Finally, conclusions and future work directions are discussed in Section 6.
 
\section{Preliminaries}
\subsection{Mathematical model of the UVMS}
In this work, the vectors are denoted with lower bold letters whereas the matrices by capital bold letters. The end effector coordinates with respect to (w.r.t) the inertial frame $\{I\}$ are denoted by $\bs{x}_e\in \mathbb{R}^6$. Let $\bs{q}=[\bs{q}^\top_a,~\bs{q}^\top_m]^\top\in \mathbb{R}^n$ be the state variables of the UVMS, where $\bs{q}_a=[\bs{\eta}_1^\top,\bs{\eta_2}^\top]^\top\in \mathbb{R}^6$ is the vector that involves the position vector  $\bs{\eta}_{1}=[x,y,z]^\top$ and  Euler-angles orientation $\bs{\eta}_{2}=[\phi,\theta,\psi]^\top$ of the vehicle w.r.t to the inertial frame $\{I\}$ and $\bs{q}_m\in \mathbb{R}^{n-6}$ is the vector of angular position of the manipulator's joints. Thus, we have \citep{Fossen2,antonelli}:
\begin{gather}
\dot{\bs{q}}_a= \bs{J}^a(\bs{q}_a)\bs{v} \label{eq1}
\end{gather}
where
\begin{align*}
\bs{J}^a(\bs{q}_a)= \begin{bmatrix}
	\bs{J}_t(\eta_2) & \bs{0}_{3 \times 3} \\
	\bs{0}_{3 \times 3} & \bs{J}_{r}(\eta_2) \\
\end{bmatrix}\in \mathbb{R}^{6\times 6}
\end{align*}
is the Jacobian matrix transforming the velocities from the body-fixed to the inertial frame
and where, $\bs{0}_{3\times 3}$ is the zero matrix of the respective dimensions, $\bs{v}$ is the vector of body velocities of the vehicle and $\bs{J}_t(\eta_2)$ and $\bs{J}_r(\eta_2)$ are the corresponding parts of the Jacobian related to position and orientation respectively. Let also $\dot{\bs{\chi}}=[\dot{\bs{\eta}}_{1}^\top,\bs{\omega}^\top]^\top$ denotes the velocity of UVMS's End-Effector, where $\dot{\bs{\eta}}_{1}$, $\bs{\omega}$ are the linear and angular velocity of the UVMS's End-Effector, respectively.  Without loss  of generality, for the augmented UVMS system we have \cite{antonelli}:
\begin{equation}
\dot{\bs{\chi}}= \bs{J}^g({\bs{q}})\boldsymbol{\zeta}\label{eq222}
\end{equation}
where $\boldsymbol{\zeta}=[\bs{v}^\top,\dot{\bs{q}}_{m,i}^\top]^\top \in \mathbb{R}^{n}$  is the velocity vector including the body velocities of the vehicle as well as the joint velocities of the manipulator and $ \bs{J}^g(\bs{q})$ is the geometric Jacobian Matrix \citep{antonelli}. In this way, the task space velocity vector of UVMS's End-Effector can be given by:
\begin{equation}
\dot{\bs{x}}_e= \bs{J}({\bs{q}})\boldsymbol{\zeta}\label{eq122}
\end{equation}
where: $\bs{J}({\bs{q}})$ is analytical Jacobian matrix given by:
\begin{gather*}
\bs{J}({\bs{q}})={\bs{J}'({\bs{q}})}^{-1}\bs{J}^g({\bs{q}})
\end{gather*}
with $\bs{J}'({\bs{q}})$ to be a Jacobian matrix that maps the Euler angle rates to angular velocities $\bs{\omega}$ and is given by:
\begin{gather*}
\bs{J}'({\bs{q}})=\begin{bmatrix}
\bs{I}_{3\times 3} &  \bs{0}_{3\times 3}\\
\bs{0}_{3\times 3} & \bs{J}''({\bs{q}})
\end{bmatrix},\\
\bs{J}''({\bs{q}})=\begin{bmatrix}
1 &  0&-\sin(\theta)\\
0&\cos(\phi)&\cos(\theta)\sin(\phi)\\
0&-\sin(\phi)&\cos(\theta)\cos(\phi)
\end{bmatrix}.
\end{gather*}

\subsection{Dynamics}
Without loss  of generality, the dynamics of the UVMS can be given as \cite{antonelli}:
\begin{gather}
\bs{M}(\bs{q})\dot{\bs{\zeta}}\!+\!\bs{C}({\bs{q}},\bs{\zeta}){\bs{\zeta}}\!+\!\bs{D}({\bs{q}},\bs{\zeta}){\bs{\zeta}}\!+\bs{g}(
\bs{q})\!+\!{\bs{J}^g}^\top\boldsymbol{\lambda}+\boldsymbol{\delta}(t)=\!\bs{\tau}\!\label{eq4}
\end{gather}
where $\boldsymbol{\delta}(t)$ are bounded disturbances including system's uncertainties as well as the external disturbances affecting on the system from the environment (sea waves and currents), $\boldsymbol{\lambda}=[\bs{f}^\top_e,\boldsymbol{\nu}^\top_e]^\top$ the generalized vector including force $\bs{f}_e$
and torque  $\boldsymbol{\nu}_e$ that the UVMS exerts on the environment at its end-effector frame. Moreover, $\bs{\tau} \in \mathbb{R}^n$ denotes the control input at the joint level, $\bs{{M}}(\bs{q})$ is the positive definite inertial matrix, $\bs{{C}}({\bs{q}},\bs{\zeta})$ represents coriolis and centrifugal terms, $\bs{{D}}({\bs{q}},\bs{\zeta})$ models dissipative effects, $\bs{{g}}(\bs{q})$ encapsulates the gravity and buoyancy effects. 
\subsection{Dynamical Systems}
Consider the initial value problem:
\begin{equation}
\dot{\xi} = H(t,\xi), \xi(0)=\xi^0\in\Omega_{\xi}, \label{eq:initial_value_problem}
\end{equation}
with $H:\mathbb{R}_{\geq 0}\times\Omega_{\xi} \to \mathbb{R}^n$, where $\Omega_{\xi}\subseteq\mathbb{R}^n$ is a non-empty open set.
\begin{definition} \citep{Sontag}
A solution $\xi(t)$ of the initial value problem \eqref{eq:initial_value_problem} is maximal if it has no proper right extension that is also a solution of \eqref{eq:initial_value_problem}. 
\end{definition}
\begin{theorem} \citep{Sontag} \label{thm:dynamical systems}
Consider the initial value problem \eqref{eq:initial_value_problem}. Assume that $H(t,\xi)$ is: a) locally Lipschitz in $\xi$ for almost all $t\in\mathbb{R}_{\geq 0}$, b) piecewise continuous in $t$ for each fixed $\xi\in\Omega_{\xi}$ and c) locally integrable in $t$ for each fixed $\xi\in\Omega_{\xi}$. Then, there exists a maximal solution $\xi(t)$ of \eqref{eq:initial_value_problem} on the time interval $[0,\tau_{\max})$, with $\tau_{\max}\in\mathbb{R}_{> 0}$ such that $\xi(t)\in\Omega_{\xi},\forall t\in[0,\tau_{\max})$.
\end{theorem}
\begin{proposition} \citep{Sontag} \label{prop:dynamical systems}
Assume that the hypotheses of Theorem \ref{thm:dynamical systems} hold. For a maximal solution $\xi(t)$ on the time interval $[0,\tau_{\max})$ with $\tau_{\max}<\infty$ and for any compact set $\Omega'_{\xi}\subseteq\Omega_{\xi}$, there exists a time instant $t'\in[0,\tau_{\max})$ such that $\xi(t')\notin\Omega'_{\xi}$.
\end{proposition}

\section{Problem Statement}
We define here the problem that we aim to solve in this paper:
\begin{problem}
Given a UVMS system as well as a desired force profile that should be applied by the UVMS on an entirely unknown model compliant environment, assuming the uncertainties on the UVMS dynamic parameters, design a feedback control law such that the following are guaranteed: 
\begin{enumerate}
	\item a predefined behavior of the system in terms of desired overshoot and prescribed transient and steady state performance.
	\item robustness with respect to the external disturbances and noise on measurement devises.
\end{enumerate}
\end{problem}

\section{Control Methodology}
In this work we assume that the UVMS is equipped with a force/torque sensor at its end-effector frame. However, we assume that its accuracy is not perfect and the system suffers from noise in the force/torque measurements. In order to combine the features of stiffness and force control, a parallel force/position regulator is designed. This can be achieved by closing a force feedback loop around a position/velocity feedback loop, since the output of the force controller becomes the reference input to the dynamic controller of the UVMS. 

\subsection{Control Design}
	Let $\bs{f}_e^d(t)$ be the desired force profile which should be exerted on the environment by the UVMS. Hence, let us define the force error:
	\begin{align} 
	\bs{e}_f(t)=\bs{f}_e(t)+\Delta\bs{f}_e(t)-\bs{f}_e^d(t)\in \mathbb{R}^3, \label{eq8}
	\end{align}
	where $\Delta\bs{f}_e(t)$ denotes the bounded noise on the force's measurement.
	Also we define the end-effector orientation error as:
	\begin{align} 
	\bs{e}_o(t)= {^o\bs{x}}_e(t)- {^o\bs{x}}^d_e(t) \in \mathbb{R}^3, \label{eq_or}
	\end{align}
	where ${^o\bs{x}}^d_e(t)\in \mathbb{R}^3$ is predefined desired orientation of the end-effector (e.g. ${^o\bs{x}}^d_e(t)=[0,~0,~0]^\top$). Now we can set the vector of desired end-effector configuration as $\bs{x}^d_e(t)= [\bs{f}_e^d(t)^\top,({^o\bs{x}^d_e(t)})^\top]^\top$. In addition the overall error vector is given as:
	\begin{align}
	\bs{e}_x(t)=[e_{x_1}(t),\ldots,e_{x_6}(t)]=[\bs{e}^\top_f(t),\bs{e}^\top_o(t)]^\top\label{eq:ov:er}
	\end{align}
	A suitable methodology for the control design in hand is that of prescribed performance control, recently proposed in \citep{Bechlioulis20141217,C-2011}, which is adapted here in order to achieve predefined transient and steady state response bounds for the errors. Prescribed performance characterizes the behavior where the aforementioned errors evolve strictly within a predefined region that is bounded by absolutely decaying functions of time, called performance functions. The mathematical expressions of prescribed performance are given by the inequalities: $-\rho_{x_j}(t)<e_{x_j}(t)<\rho_{x_j}(t),~ j=1,\ldots,6$, where $\rho_{x_j}:[t_0,\infty)\rightarrow\mathbb{R}_{>0}$ with $\rho_{x_j}(t)=(\rho^0_{x_j}-\rho_{x_j}^\infty)e^{-l_{x_j}t}+\rho_{x_j}^\infty$ and $l_{x_j}>0,\rho^0_{x_j}>\rho^\infty_{x_j}>0$, are designer specified, smooth, bounded and decreasing positive functions of time with positive parameters $l_{x_j},\rho^\infty_{x_j}$, incorporating the desired transient and steady state performance respectively. In particular, the decreasing rate of $\rho_{x_j}$, which is affected by the constant $l_{x_j}$ introduces a lower bound on the speed of convergence of $e_{x_j}$. Furthermore, the constants $\rho^\infty_{x_j}$ can be set arbitrarily small, achieving thus practical convergence of the errors to zeros.
	
	Now, we propose a state feedback control protocol $\boldsymbol{\tau}(t)$, that does not incorporate any information regarding the UVMS dynamic model \eqref{eq4} and model of complaint environment, and achieves tracking of the smooth and bounded desired force trajectory $\bs{f}_e^d(t)\in \mathbb{R}^3$ as well as ${^o\bs{x}}^d_e(t)$ with an priori specified convergence rate and steady state error. Thus, given the errors \eqref{eq:ov:er}:

\textbf{Step I-a}: Select the corresponding functions $\rho_{x_j}(t)=(\rho^0_{x_j}-\rho_{x_j}^\infty)e^{-l_{x_j}t}+\rho_{x_j}^\infty$ with $\rho^0_{x_j}>|e_{x_j}(t_0)|, \forall j\in\{1\ldots,6\}$ $\rho^0_{x_j}>\rho^\infty_{x_j}>0$, $ l_{x_j}>0,\forall j\in\{1,\ldots 6\}$, in order to incorporate the desired transient and steady state performance specification and define the normalized errors:
\begin{align}
\xi_{x_j}(t)=\frac{e_{x_j}(t)}{\rho_{x_j}(t)},~j=\{1,\ldots,6\}\label{eq11}
\end{align}
\textbf{Step I-b}: Define the transformed errors $\varepsilon_{x_j}$ as:
\begin{align}
\varepsilon_{x_j}(\xi_{x_j})=\ln \Big(\frac{1+\xi_{x_j}}{1-\xi_{x_j}}\Big),~j=\{1,\ldots,6\}\label{eq12}
\end{align}
Now, the reference velocity as $\dot{\bs{x}}^r_e=[\dot{x}^r_{e_1},\ldots,\dot{x}^r_{e_6}]^\top$ is designed as:
\begin{align}
\dot{x}^r_{e_j}(t)=-k_{x_j}\varepsilon_{x_j}(\xi_{x_j}),~k_j>0,~j=\{1,\ldots,6\}\label{eq13}
\end{align}
The task-space desired motion profile $\dot{\bs{x}}^r_e$ can be extended to the joint level using the kinematic equation \eqref{eq122}:
\begin{equation}
{\bs{\zeta}}^r(t)=\bs{J}(\bs{q})^{+}\dot{\bs{x}}^r_e  \label{eq9}
\end{equation}
where $\bs{J}(\bs{q})^{+}$ denotes the Moore-Penrose pseudo-inverse of Jacobian $\bs{J}(\bs{q})$.
\begin{remark}
	It is worth mentioning that the $\dot{\bs{x}}^r_e$ can also be extended to the joint level via:  
	\begin{align*}
	{\bs{\zeta}}^r(t)=\bs{J}(\bs{q})^{\#}\dot{\bs{x}}^r_e+\big(\bs{I}_{n\times n}\!-\!\bs{J}(\bs{q})^{\#}\bs{J}\big(\bs{q}\big)\big)\dot{\bs{x}}^0 
	\end{align*}
	where $\bs{J}(\bs{q})^{\#}$ denotes the generalized pseudo-inverse \citep{citeulike:6536020} of Jacobian $\bs{J}(\bs{q})$ and $\dot{\bs{x}}^0$ denotes secondary tasks which can be regulated independently to achieve secondary goals (e.g., maintaining manipulator's joint limits, increasing of manipulability) and does not contribute to the end effector's velocity \citep{Simetti2016877}. 
\end{remark}

\textbf{Step II-a}: Define the velocity error vector at the end-effector frame as:
\begin{align}
\bs{e}_\zeta(t)=[{e}_{\zeta_1}(t),\ldots,{e}_{\zeta_n}(t)]^\top= {{\bs{\zeta}}}(t)- {{\bs{\zeta}}}^r(t) \label{eq14}
\end{align}
and select the corresponding functions $\rho_{\zeta_j}(t)=(\rho^0_{\zeta_j}-\rho_{\zeta_j}^\infty)e^{-l_{\zeta_j}t}+\rho_{\zeta_j}^\infty$ with $\rho^0_{\zeta_j}>|e_{\zeta_j}(t_0)|,\forall j\in\{1\ldots,n\}$, $\rho^0_{\zeta_j}>\rho^\infty_{\zeta_j}>0$, $ l_{\zeta_j}>0,\forall j\in\{1,\ldots n\}$, and define the normalized velocity errors $\boldsymbol{\xi}_\zeta$ as:
\begin{align}
\boldsymbol{\xi}_{\zeta}(t)=[\xi_{\zeta_1},\ldots,\xi_{\zeta_n}]^\top=\bs{P}^{-1}_\zeta(t)\bs{e}_\zeta(t)\label{eq15}
\end{align}
where $\bs{P}_\zeta(t)=\text{diag}\{\rho_{\zeta_j}\},j\in\{1,\ldots,n\}$.\\
\textbf{Step II-b}:  Define the transformed errors $\boldsymbol{\varepsilon}_{\zeta}(\boldsymbol{\xi}_{\zeta})=[\varepsilon_{\zeta_1}(\xi_{\zeta_1}),\ldots,\varepsilon_{\zeta_n}(\xi_{\zeta_n})]^\top$ and the signal $\bs{R}_{\zeta}(\boldsymbol{\xi}_{\zeta})=\text{diag}\{r_{\zeta_j}\}$, $~j\in\{1,\ldots,n\}$ as:
\begin{align}\label{eq16}
&\boldsymbol{\varepsilon}_{\zeta}(\boldsymbol{\xi}_{\zeta})=\Big[\ln \Big(\frac{1+\xi_{\zeta_1}}{1-\xi_{\zeta_1}}\Big),\ldots,\ln \Big(\frac{1+\xi_{\zeta_n}}{1-\xi_{\zeta_n}}\Big)\Big]^\top\\
& \bs{R}_{\zeta}(\boldsymbol{\xi}_{\zeta})\!=\!\text{diag}\{r_{\zeta_j}\!(\xi_{\zeta_j}\!)\}\!=\!\text{diag}\!\Big\{\!\frac{2}{1-\xi_{\zeta_j}^2\!}\Big\},j\!=\!\{1,\ldots,n\}\label{eq17}
\end{align}
and finally design the state feedback control law $\tau_j,~j\in\{1,\ldots,n\}$ as:
\begin{align}
\tau_j\!(\xi_{x_j}\!,\xi_{\zeta_j}\!,t)=-k_{\zeta_j}\frac{r_{\zeta_j}(\xi_{\zeta_j})\varepsilon_{\zeta_j}(\xi_{\zeta_j})}{\rho_{\zeta_j}(t)},~j=\{1,\ldots,n\}\label{eq18}
\end{align}
where $k_{\zeta_j}$ to be a positive gain. The control law \eqref{eq18} can be written in vector form as:
\begin{align}
\bs{\tau}(\bs{e}_x(t),\bs{e}_\zeta(t),t)&=[ \tau_1(\xi_{x_1},\xi_{\zeta_1},t),\ldots, \tau_n(\xi_{x_n},\xi_{\zeta_n},t)]^\top\nonumber\\
&-\bs{K}_\zeta\bs{P}^{-1}(t)\bs{R}_{\zeta}(\boldsymbol{\xi}_{\zeta})\boldsymbol{\varepsilon}_{\zeta}(\boldsymbol{\xi}_{\zeta})\label{eq19}
\end{align}
with $\bs{K}_\zeta$ to be the diagonal matrix containing $k_{\zeta_j}$. Now we are ready to state the main theorem of the paper:
\begin{theorem}
Given the error defined in \eqref{eq:ov:er} and the required transient and steady state performance specifications, select the exponentially decaying performance function $\rho_{x_j}(t)$, $\rho_{\zeta_j}(t)$ such that the desired performance specifications are met. Then the state feedback control law of \eqref{eq19} guarantees tracking of the trajectory $\bs{f}_e^d(t)\in \mathbb{R}^3$ as well as ${^o\bs{x}}^d_e(t)$:
\begin{align*}
\lim_{t\rightarrow\infty}\bs{f}_e(t)=\bs{f}^d_e(t)~\text{and}\lim_{t\rightarrow\infty}{^o\bs{x}_e(t)}={^o\bs{x}^d_e(t)}
\end{align*}
 with the desired transient and steady state performance specifications.
\end{theorem}
\begin{pf}
For the proof we follow parts of the approach in \citep{Bechlioulis20141217}. We start by differentiating \eqref{eq11} and \eqref{eq15} with respect to the time and substituting the system dynamics \eqref{eq4} as well as \eqref{eq13} and \eqref{eq18} and employing \eqref{eq:ov:er} and \eqref{eq14}, obtaining:
\begin{align}
\dot{\xi}_{x_j}(\xi_{x_j},t)&=h_{x_j}(\xi_{x_j},t)\nonumber\\
&=\rho^{-1}_{x_j}(t)(\dot{e}_{x_j}(t)-\dot{\rho}_{x_j}\!(t)\xi_{x_j} )\nonumber\\
&=\rho^{-1}_{x_j}(t)(-k_{x_j}\varepsilon_{x_j}(\xi_{x_j})+\bs{J}_{(j,:)}\bs{P}_\zeta\bs{\xi}_\zeta-\dot{x}^d_{e_j}(t))\nonumber\\
&-\rho^{-1}_{x_j}(t)(\dot{\rho}_{x_j}\!(t)\xi_{x_j}), \forall j\in\{1,\ldots,6\}\label{eq21}
\end{align}
\begin{align}
\dot{\boldsymbol{{\xi}}}_{\zeta}(\xi_{\zeta},t&)=h_{\zeta}(\bs{\xi}_{\zeta},t)\nonumber\\
& = \bs{P}_\zeta^{-1}(\dot{\bs{\zeta}}-\dot{\bs{\zeta}}^r)-\dot{\bs{P}}_\zeta^{-1}{\boldsymbol{\xi}}_\zeta ) \nonumber\\
& = -\bs{K}_\zeta\bs{P}_\zeta^{-1}\bs{M}^{-1}\bs{P}_\zeta^{-1}\bs{R}_\zeta\boldsymbol{\varepsilon}_{\zeta}-\nonumber\\
&-\bs{P}_\zeta^{-1}\Big[\bs{M}^{-1}\Big(\bs{C}\cdot(\bs{P}_\zeta\boldsymbol{\xi}_\zeta+{\bs{\zeta}}^r)+\bs{D}\cdot(\bs{P}_\zeta\boldsymbol{\xi}_\zeta+{\bs{\zeta}}^r)\nonumber\\
&+\bs{g}+{\bs{J}^g}^\top\boldsymbol{\lambda}+\!\boldsymbol{\delta}(t)\Big)+\dot{\bs{P}}_\zeta\boldsymbol{\xi}_\zeta+\frac{\partial}{\partial t}{\bs{\zeta}}^r\Big]\label{eq22}
\end{align}
where $\bs{J}_{(j,:)}$ denotes all elements of jacobian $\bs{J}$ at its $j$ row. Now let us to define the vectors of normalized state error and the generalized normalized error as $\boldsymbol{\xi}_{x}=[{\xi}_{x_1},\ldots,{\xi}_{x_6}]^\top\!$, and $\boldsymbol{\xi}=[\boldsymbol{\xi}_x^\top,\boldsymbol{\xi}_\zeta^\top]^\top$, respectively. Moreover, let us define $\dot{\bs{\xi}}_x=h_x(\bs{\xi_x},t)=[h_{x_1}(\xi_{x_1},t),\ldots,h_{x_6}(\xi_{x_6},t)]^\top$.  The equations of \eqref{eq21} and \eqref{eq22} now can be written in compact form as:
\begin{align}
\dot{\boldsymbol{\xi}}=h(\boldsymbol{\xi},t)= [h_x^\top(\bs{\xi_x},t),h_\zeta^\top(\bs{\xi_\zeta},t)]\top\label{eq23}
\end{align}
Let us define the open set $\Omega_\xi=\Omega_{\xi_x}\times\Omega_{\xi_\zeta}$ with $\Omega_{\xi_x}=(-1,1)^6$ and $\Omega_{\xi_\zeta}=(-1,1)^n$. In what follows, we proceed in two phases. First we ensure the existence of a unique maximal solution $\boldsymbol{\xi}(t)$ of \eqref{eq23} over the set $\Omega_\xi$ for a time interval $[0,t_{\text{max}}]$ (i.e., $\boldsymbol{\xi}(t)\in\Omega_\xi, \forall t\in[0,t_{\text{max}}]$). Then, we prove that the proposed controller \eqref{eq19} guarantees, for all $t\in[0,t_{\text{max}}]$ the boundedness of all closed loop signal of \eqref{eq23} as well as that $\boldsymbol{\xi}(t)$ remains strictly within the set $\Omega_\xi$, which leads that $t_{\text{max}}=\infty$ completes the proof.  

\textbf{Phase A}: The set $\Omega_\xi$ is nonempty and open, thus by selecting $\rho^0_{x_j}>|e_{x_j}(0)|$, $\forall j\in\{1,\ldots 6\}$ and $\rho^0_{v_j}>|e_{v_j}(0)|$, $\forall j\in\{1,\ldots n\}$ we guarantee that $\boldsymbol{\xi}_x(0)\in\Omega_{\xi_x}$ and $\boldsymbol{\xi}_\zeta(0)\in\Omega_{\xi_\zeta}$. Additionally, $h$  is continuous on $t$ and locally Lipschitz on $\boldsymbol{\xi}$ over $\Omega_\xi$.  Therefore, the hypotheses of Theorem\ref{thm:dynamical systems} hold and the existence of a maximal solution $\boldsymbol{\xi}(t)$ of \eqref{eq23} on a time interval $[0,t_{\text{max}}]$ such that $\boldsymbol{\xi}(t) \in \Omega_\xi,~\forall t\in[0,t_{\text{max}}]$ is ensured.
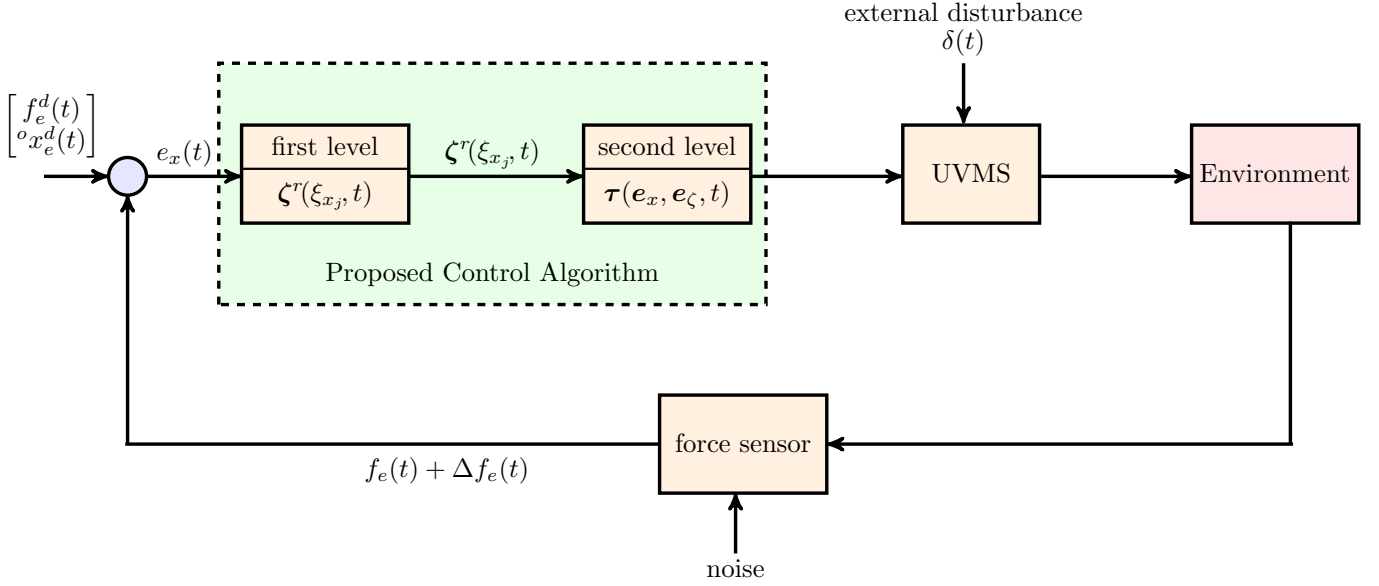
\begin{figure*}
	\centering
	\begin{tikzpicture}
	
	\node at (-13.0, 2.87) {$\begin{bmatrix}f_e^d(t)\\ ^o{x}_e^d(t) \end{bmatrix}$};
	
	\filldraw[fill=green!10, dashed, line width=.045cm] (-10.80, 0.5) rectangle +(7.2, 3.2);
	
	\node at (-7.2, 0.9) {$\text{Proposed Control Algorithm}$};
	
	\filldraw[fill=blue!10, line width=.045cm]  (-12.0,2.20) circle (0.25cm);
	\node at (-11.25,2.50) {$e_x(t)$};
	\draw [color=black,thick,->,>=stealth', line width=.045cm](-11.75, 2.2) to (-10.5, 2.2);
	\draw [color=black,thick,->,>=stealth', line width=.045cm](-13.1, 2.2) to (-12.25, 2.2);
	
	\filldraw[fill=orange!12, line width=.045cm] (-10.5, 1.58) rectangle +(2.2, 1.3);
	\node at (-9.4,2.60) {$\text{first level}$};
	\node at (-9.4,1.95) {${\bs{\zeta}}^r\!(\xi_{x_j}\!,t)$};
	\draw [black, line width = 0.030cm] (-10.5, 2.30) -- (-8.3,2.30);
	\draw [color=black,thick,->,>=stealth', line width=.045cm](-8.3, 2.2) to (-6.0, 2.2);
	\node at (-7.2,2.50) {${\bs{\zeta}}^r\!(\xi_{x_j}\!,t)$};
	
	\filldraw[fill=orange!12, line width=.045cm] (-6.0, 1.58) rectangle +(2.2, 1.3);
	\node at (-4.9,2.60) {$\text{second level}$};
	\node at (-4.9,1.95) {$\bs{\tau}(\bs{e}_x,\bs{e}_\zeta,t)$};
	\draw [black, line width = 0.030cm] (-6.00, 2.30) -- (-3.8,2.30);
	\draw [color=black,thick,->,>=stealth', line width=.045cm](-3.8, 2.2) to (-1.8, 2.2);
	
	\filldraw[fill=orange!12, line width=.045cm] (-1.8, 1.58) rectangle +(1.80, 1.3);
	\node at (-0.9,2.25) {$\text{UVMS}$};
	\draw [color=black,thick,->,>=stealth', line width=.045cm](-1.0, 3.7) to (-1.0, 2.90);
	\node at (-1.0,4.38) {$\text{external disturbance}$};
	\node at (-1.0,3.97) {$\delta(t)$};
	\draw [color=black,thick,->,>=stealth', line width=.045cm](0.0, 2.2) to (2.0, 2.2);
	
	\filldraw[fill=red!10, line width=.045cm] (2.0, 1.58) rectangle +(2.2, 1.3);
	\node at (3.1,2.25) {$\text{Environment}$};
	
	\filldraw[fill=orange!12, line width=.045cm] (-5.0, -2.00) rectangle +(2.20, 1.3);
	\node at (-3.9,-1.33) {$\text{force sensor}$};
	\draw [color=black,thick,->,>=stealth', line width=.045cm](-4.0, -2.8) to (-4.0, -2.0);
	\node at (-4.0,-3.00) {$\text{noise}$};
	\draw [black, line width = .045cm] (-12.03,-1.35) -- (-5.0,-1.350);
	\draw [color=black,thick,->,>=stealth', line width=.045cm](-12.0, -1.35) to (-12.0, 1.95);
	\draw [black, thick,->,>=stealth', line width = .045cm] (3.335,-1.350) to (-2.8,-1.35);
	\draw [black, line width = .045cm] (3.3,-1.35) -- (3.3,1.6);
	\node at (-7.8,-1.70) {$f_e(t)+\Delta f_e(t)$};	
	\end{tikzpicture}
	\centering
	\caption{The closed loop block diagram of the proposed control scheme.}
	\label{fig:closed_loop_control_scheme}
\end{figure*}

\textbf{Phase B}: In the Phase A we have proven that $\boldsymbol{\xi}(t) \in \Omega_\xi,~\forall t\in[0,t_{\text{max}}]$, thus it can be concluded that:
\begin{subequations}\label{eq24}
	\begin{gather}
\xi_{x_j}(t)=\frac{e_{x_j}}{\rho_{x_j}}\in(-1,1),~ \forall j\{1,\ldots,6\} \\
\xi_{\zeta_j}(t)=\frac{e_{\zeta_j}}{\rho_{\zeta_j}}\in(-1,1),~ \forall j\{1,\ldots,n\}
\end{gather}
\end{subequations}
for all $t\in[0,t_{\text{max}}]$, from which we obtain that $e_{x_j}(t)$ and $e_{\zeta_j}(t)$ are absolutely bounded by $\rho_{x_j}$ and $\rho_{\zeta_j}$, respectively. Therefore, the error vectors $\varepsilon_{x_j}(\xi_{x_j}),\forall j\in\{1,\ldots,6\}$ and $\varepsilon_{\zeta_j}(\xi_{\zeta_j}),\forall j\in\{1,\ldots,n\}$ defined in \eqref{eq12} and \eqref{eq16}, respectively, are well defined for all $t\in [0,t_{\text{max}}]$. Hence, consider the positive definite and radially unbounded functions $V_{x_j}(\varepsilon_{x_j})=\varepsilon^2_{x_j},~\forall j\{1,\ldots,6\}$. Differentiating of $V_{x_j}$ w.r.t time and substituting \eqref{eq21}, results in:
\begin{align}
\dot{V}_{x_j}\!=-\frac{4\varepsilon_{x_j}\rho^{-1}_{x_j}}{(1-\xi^2_{x_j})}\!\Big(k_{x_j}\varepsilon_{x_j}(\xi_{x_j})\!+\!\dot{x}^d_{e_j}\!+\!\dot{\rho}_{x_j}\!(t)\xi_{x_j}\!-\!\bs{J}_{(j,:)}\bs{P}_\zeta\bs{\xi}_\zeta\! \Big)\label{eq25}
\end{align}
It is well known that the Jacobian $\bs{J}$ is depended only on bounded vehicle's orientation and angular position of manipulator's joint. Moreover, since,  $\dot{x}^d_{e_j}$, $\rho_{x_j}$ and $\rho_{v_j}$ are bounded by construction and $\xi_{x_j}$,$\xi_{v_j}$ are also bounded in $(-1,1)$, owing to \eqref{eq24},  $\dot{V}_{x_j}$ becomes:
\begin{align}
\dot{V}_{x_j} \leq\frac{4\rho^{-1}_{x_j}}{(1-\xi^2_{x_j})}\!\Big(B_x |\varepsilon_{x_j}| -k_{x_j}|\varepsilon_{x_j}|^2\Big)\label{eq26}
\end{align}
$\forall t\in [0,t_{\text{max}}]$, where $B_x$ is an unknown positive constant independent of $t_{\text{max}}$ satisfying $B_x>|\dot{x}^d_{e_j}\!+\!\dot{\rho}_{x_j}\!(t)\xi_{x_j}\!-\!\bs{J}_{(j,:)}\bs{P}_\zeta\bs{\xi}_\zeta|$. Therefore, we conclude that $\dot{V}_{x_j}$ is negative when $\varepsilon_{x_j}>\frac{B_x}{k_{j_x}}$ and subsequently that
\begin{align}
|\varepsilon_{x_j}(\xi_{x_j}(t)) |\leq \bar{\varepsilon}_{x_j}=\max\{\varepsilon_{x_j}(\xi_{x_j}(0)),\frac{B_x}{k_{j_x}}\}\label{eq27}
\end{align}
$\forall t\in [0,t_{\text{max}}], \forall j\{1,\ldots,6\}$. Furthermore, from \eqref{eq12}, taking the inverse logarithm, we obtain:
\begin{align}
-1<\frac{e^{-\bar{\varepsilon}_{x_j}}-1 }{e^{-\bar{\varepsilon}_{x_j}}+1}=\underline{\xi}_{x_j}  \leq  \xi_{x_j}(t)\leq\bar{\xi}_{x_j} =\frac{e^{\bar{\varepsilon}_{x_j}}-1 }{e^{\bar{\varepsilon}_{x_j}}+1}<1   \label{eq28}
\end{align}
$\forall t\in [0,t_{\text{max}}],~j\in\{1,\ldots,6\}$. Due to \eqref{eq28}, the reference velocity vector $\dot{\bs{x}}^r_e$ as defined in \eqref{eq13}, remains bounded for all $t\in[0,t_{\text{max}}]$. Moreover, invoking $\dot{\bs{x}}_e={\dot{\bs{x}}}^r_e(t)+\bs{P}_v(t)\bs{\xi}_v$ from \eqref{eq14}, \eqref{eq15} and \eqref{eq24}, we also conclude the boundedness of $\dot{\bs{x}}_e$ for all $t\in [0,t_{\text{max}}]$. Finally, differentiating ${\dot{\bs{x}}}^r_e(t)$ w.r.t time and employing \eqref{eq21}, \eqref{eq24} and \eqref{eq28}, we conclude the boundedness of $\frac{\partial}{\partial t}{\dot{\bs{x}}}^r_e(t)$, $\forall t\in [0,t_{\text{max}}]$.
Applying the aforementioned line of proof, we consider the positive definite and radially unbounded function $V_\zeta(\bs{\varepsilon}_\zeta)=\frac{1}{2}||\bs{\varepsilon}_\zeta||^2$. By differentiating $V_\zeta$ with respect to time,  substituting \eqref{eq22} and by employing continuity of $\bs{M}$, $\bs{C}$, $\bs{D}$, $\bs{g}$, $\boldsymbol{\lambda}$, $\boldsymbol{\delta}$, $\boldsymbol{\xi}_x$, $\boldsymbol{\xi}_\zeta$,$\dot{\bs{P}}_\zeta$, $\frac{\partial}{\partial t}{{\bs{\zeta}}}^r$, $\forall t\in [0,t_{\text{max}}]$, we obtain:
\begin{align*}
\dot{V}_\zeta\leq ||\bs{P}^{-1}_\zeta\bs{R}_\zeta(\boldsymbol{\xi}_\zeta)\boldsymbol{\varepsilon}_\zeta||\Big(B_\zeta-\bs{K}_\zeta\lambda_M||\bs{P}^{-1}_\zeta\bs{R}_\zeta(\boldsymbol{\xi}_\zeta) \boldsymbol{\varepsilon}_\zeta||   \Big)
\end{align*}
$\forall t\in [0,t_{\text{max}}]$, where $\lambda_M$ is the minimum singular value of the positive definite matrix $\bs{M}^{-1}$ and $B_v$ is a positive constant independent of $t_{\text{max}}$, satisfying 
\begin{align*}
B_\zeta\geq &|| \bs{M}^{-1}\Big( \bs{C}\cdot(\bs{P}_\zeta\boldsymbol{\xi}_\zeta +{\bs{\zeta}}^r(t)) + \bs{D}\cdot(\bs{P}_\zeta\boldsymbol{\xi}_\zeta +{\bs{\zeta}}^r(t))  \\
&+\bs{g}+{\bs{J}^g}^\top\boldsymbol{\lambda}+\boldsymbol{\delta}(t)+ \dot{\bs{P}}_\zeta\boldsymbol{\xi}_\zeta+\frac{\partial}{\partial t}{{\bs{\zeta}}}^r  \Big)   ||
\end{align*}
Thus, $\dot{V}_\zeta$ is negative when $||\bs{P}^{-1}_\zeta\bs{R}_\zeta(\boldsymbol{\xi}_\zeta) \boldsymbol{\varepsilon}_\zeta|| >B_\zeta(\bs{K}_\zeta\lambda_M)^{-1}$, which by employing the definitions of $\bs{P}_\zeta$ and  $\bs{R}_\zeta$, becomes $||\boldsymbol{\varepsilon}_\zeta ||> B_\zeta(\bs{K}_\zeta\lambda_M)^{-1}\max\{\rho^0_{\zeta_1},\ldots,\rho^0_{\zeta_n}\} $. Therefore, we conclude that:\begin{small}
\begin{align*}
||\boldsymbol{\varepsilon}_\zeta (\boldsymbol{\xi}_\zeta\!(\!t\!))\!\leq\boldsymbol{\bar{\varepsilon}}_\zeta=\!\max\!\Big\{\!\boldsymbol{\varepsilon}_\zeta (\boldsymbol{\xi}_\zeta(0)),\! B_\zeta(\!\bs{K}_\zeta\lambda_M)^{-1}\cdot\!\max\{\rho^0_{\zeta_1},\ldots,\rho^0_{\zeta_n}\}\!  \Big\}
\end{align*}\end{small}
$\forall t\in [0,t_{\text{max}}]$. Furthermore, from \eqref{eq17}, invoking that $|\varepsilon_{\zeta_j}|\leq || \boldsymbol{\varepsilon}_\zeta||$, we obtain:
\begin{align}
-1<\frac{e^{-\bar{\varepsilon}_{\zeta_j}}-1 }{e^{-\bar{\varepsilon}_{\zeta_j}}+1}=\underline{\xi}_{\zeta_j}  \leq  \xi_{\zeta_j}(t)\leq\bar{\xi}_{\zeta_j} =\frac{e^{\bar{\varepsilon}_{\zeta_j}}-1 }{e^{\bar{\varepsilon}_{\zeta_j}}+1}<1   \label{eq29}
\end{align}
$\forall t\in [0,t_{\text{max}}],~j\in\{1,\ldots,n\}$ which also leads to the boundedness of the control law \eqref{eq19}. Now, we will show that the $t_{\text{max}}$ can be extended to $\infty$. Obviously, notice by \eqref{eq28} and \eqref{eq29} that $\boldsymbol{\xi}(t)\in\Omega^{'}_\xi=\Omega^{'}_{\xi_x}\times \Omega^{'}_{\xi_\zeta},\forall t\in [0,t_{\text{max}}]$, where:
\begin{align*}
 \Omega^{'}_{\xi_x}=[\underline{\xi}_{x_1},\bar{\xi}_{x_1}]\times\ldots,\times[\underline{\xi}_{x_6},\bar{\xi}_{x_6}  ],\\
  \Omega^{'}_{\xi_\zeta}=[\underline{\xi}_{\zeta_1},\bar{\xi}_{\zeta_1}]\times\ldots,\times[\underline{\xi}_{\zeta_n},\bar{\xi}_{\zeta_n}  ],
\end{align*}
are nonempty and compact subsets of $\Omega_{\xi_x}$ and $\Omega_{\xi_\zeta}$, respectively. Hence, assuming that $t_{\text{max}}<\infty$ and since $\Omega_\xi\subset \Omega^{'}_\xi$, Proposition 1, dictates the existence of a time instant $t{^{'}}\in \forall t\in [0,t_{\text{max}}]$ such that $\boldsymbol{\xi}(t^{'})\notin \Omega^{'}_\xi$, which is a clear contradiction. Therefore, $t_{\text{max}}=\infty$. Thus, all closed loop signals remain bounded and moreover $\boldsymbol{\xi}(t)\in \Omega^{'}_\xi,\forall t\geq0$.
Finally, from \eqref{eq11} and \eqref{eq28} we conclude that:
\begin{align}
-\rho_{x_j}<\frac{e^{-\bar{\varepsilon}_{x_j}}-1 }{e^{-\bar{\varepsilon}_{x_j}}+1}\rho_{x_j} \leq  e_{x_j}(t)\leq \rho_{x_j}\frac{e^{\bar{\varepsilon}_{x_j}}-1 }{e^{\bar{\varepsilon}_{x_j}}+1}<\rho_{x_j}   \label{eq30}
\end{align}
for $j\in\{1,\ldots,6\}$ and for all $t\geq 0$ and consequently, completes the proof. 
\end{pf}
\begin{remark}
From the aforementioned proof, it is worth noticing that the proposed control scheme is model free with respect to the matrices $\bs{M}$, $\bs{C}$, $\bs{D}$, $\bs{g}$ as well as the external disturbances $\boldsymbol{\delta}$ that affect only the size of $\bar{\varepsilon}_{x_j}$ and of  $\bar{\varepsilon}_{\zeta_j}$,  but leave unaltered the achieved convergence properties as \eqref{eq30} dictates. In fact, the actual transient and steady state performance is determined by the selection of the performance function $\rho_{c_j},c\in \{x,\zeta\}$.  Finally the closed loop block diagram of the proposed control scheme is indicated in Fig.\ref{fig:closed_loop_control_scheme}.
\end{remark}

\section{Simulation Results}
Simulation studies were conducted employing a hydrodynamic simulator built in MATLAB. The dynamic equation of UVMS used in this simulator is derived following \cite{Schjølberg94modellingand}. The UVMS model considered in the simulations is the SeaBotix LBV150 equipped with a small 4 DoF manipulator. We consider a scenario involving 3D motion in workspace, where the end-effector of the UVMS is in interaction on a compliant environment with stiffens matrix  $\bs{K}_f=\text{diag}\{2\}$ which is unknown for the controller.
\begin{figure}[h!]
	\centering
	\includegraphics[scale = 0.30]{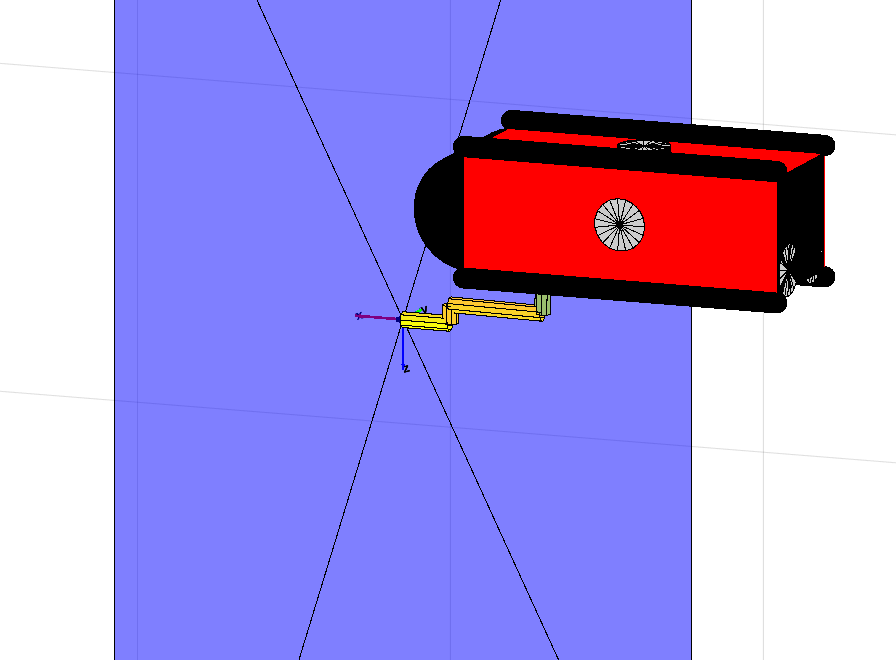}
	\caption{Workspace including the UVMS and the compliant environment. The UVMS is run under the influence of external disturbances.}
	\label{fig:workspace}       
\end{figure}
The workspace at the initial time, including UVMS and the compliant environment are depicted in Fig.\ref{fig:workspace}. More specifically, we adopt: $\bs{f}_e(0)=[0,0,0]^\top$ and $^o\bs{x}_e=[0.2,0.2,-0.2]^\top$. It means that at the initial time of the simulation study we assume that the uvms has attached the compliant environment with a rotation at its end-effector frame. The control gains for the two set of the simulation studies were selected as follows: $k_{x_j}=-0.2 j\in\{1,\ldots,6\}$, $k_{v_j}=-5 j\in\{1,\ldots,n\}$. Moreover, the dynamic parameters of UVMSs as well as the stiffens matrix $\bs{K}_f$ were considered unknown for the controller. The parameters of the performance functions in sequel stimulation studies were chosen as follows:$\rho^0_{x_j}=1,~j\in\{1,2,3\}$, $\rho^0_{x_j}=0.9,~j\in\{4,5,6\}$, $\rho^0_{v_j}=1,~j\in\{1,\ldots,n\}$, $\rho^\infty_{x_j}=0.2~j\in\{1,\ldots,6\}$,  $\rho^\infty_{v_j}=0.2~j\in\{1,\ldots,6\}$,    $\rho^\infty_{v_j}=0.4~j\in\{7,\ldots,n\}$, $l_{x_j}=3~j\in\{1,2,3\}$, $l_{v_j}=2.2~j\in\{1,\ldots,n\}$. Finally, the whole system was running under the influence of external disturbances (e.g., sea current) acting along $x$, $y$ and $z$ axes (on the vehicle body) given by $0.15\sin(\frac{2\pi}{7}t)$, in order to test the robustness of the proposed scheme. Moreover, bounded noise on measurement devices were considered during the simulation study. In the the presented simulation scenario,  a desired force trajectory should be exerted to the environment while predefined orientation $^o\bs{x}^d_e=[0.0,0.0,0.0]^\top$ must be kept. One should bear in mind that this is a challenging task owing to dynamic nature of the underwater environment. The UVMS's model uncertainties, noise of measurement devices as well as external disturbances in this case can easily cause unpredicted instabilities to the system. The desired force trajectory which should be exerted by UVMS is defined as $f^d_{e_1}=0.4\sin(\frac{2\pi}{2}t)+0.4$. The results are depicted in Figs 3-5. Fig\ref{fig:forc} show the evolution of the force trajectory. Obviously, the actual force exerted by the UVMS (indicated by red color) converges to the desired one (indicated by green color) without overshooting and follows the desired force profile. The evolution of the errors at the first and second level of the proposed controller are indicated in Fig.\ref{fig:ppx_force} and Fig.\ref{fig:ppv_force}, respectively. It can be concluded that even with the influence of external disturbances as well as noise in measurements, the errors in all directions converge close to zero and remain bounded by the performance functions.
\begin{figure}[h!]
	\centering
	\includegraphics[scale = 0.35]{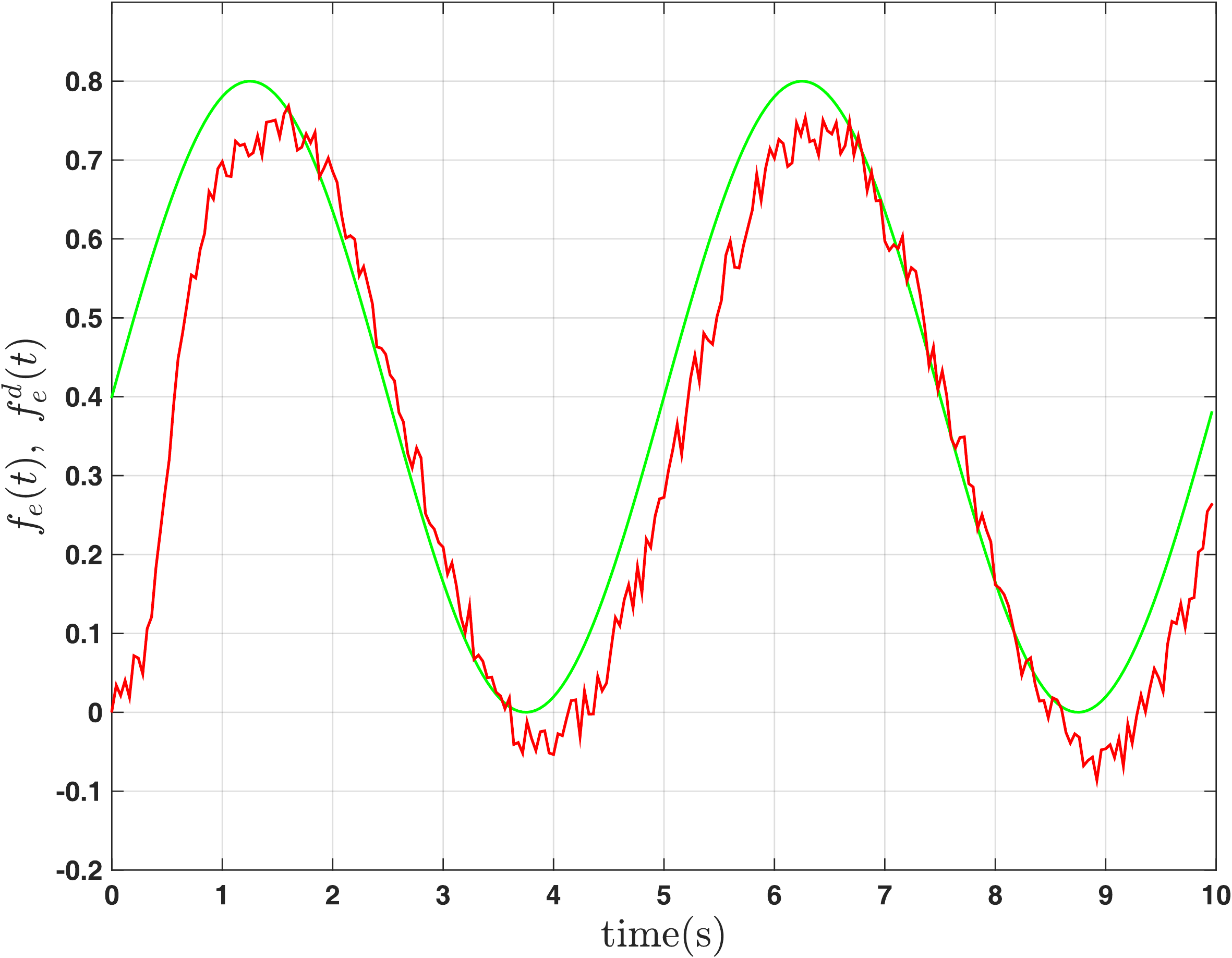}
	\caption{Trajectory scenario: The evolution of the force trajectory. The desired force trajectory and the actual force exerted by the UVMS are indicated by green and red color respectively.}
	\label{fig:forc}       
\end{figure}
\begin{figure}[h!]
	\centering
	\includegraphics[scale = 0.56]{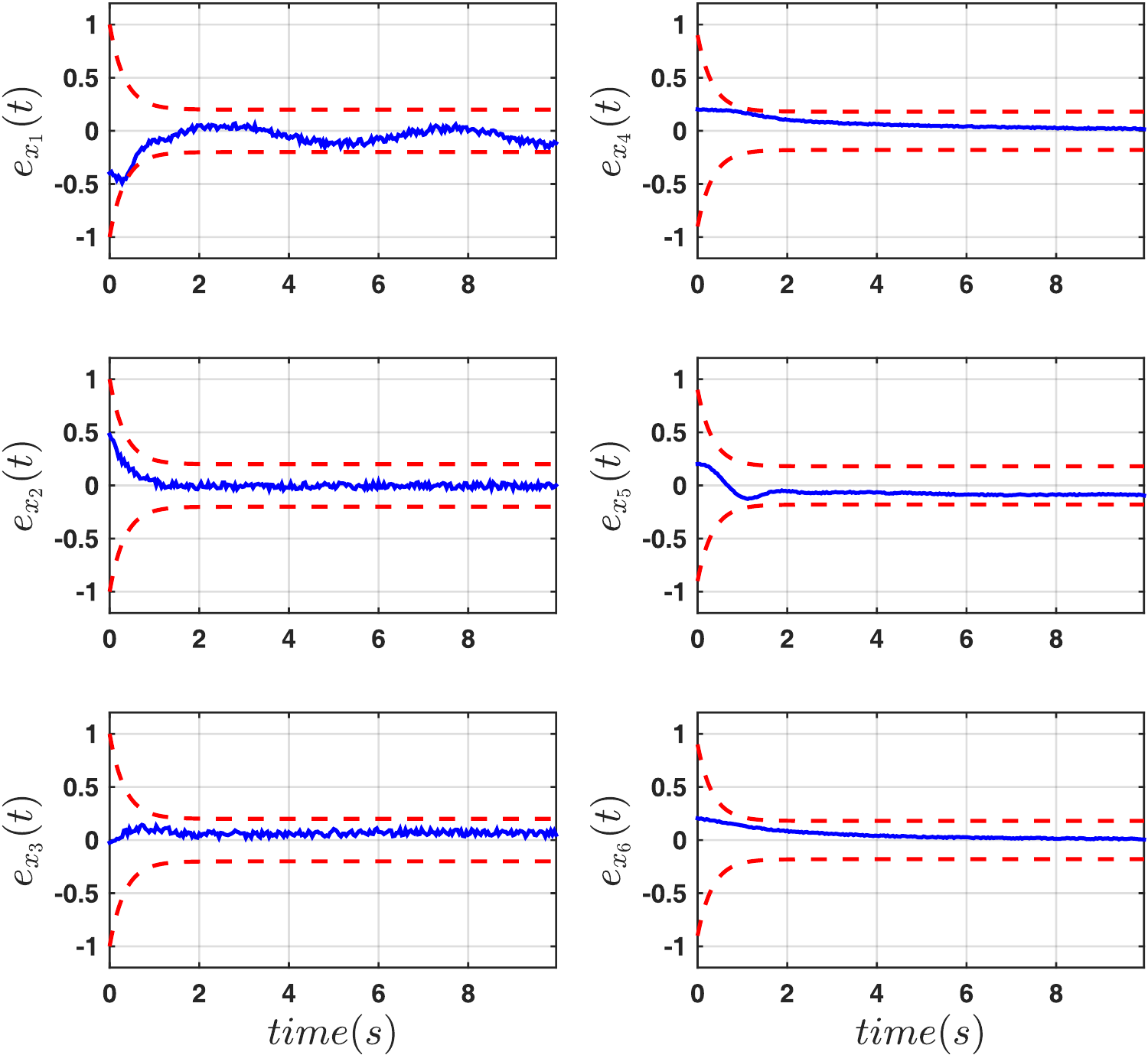}
	\caption{Trajectory scenario: The evolution of the errors at the first level of the proposed control scheme. The errors and performance bounds are indicated by blue and red color respectively.}
	\label{fig:ppx_force}       
\end{figure}
\begin{figure}[h!]
	\centering
	\includegraphics[scale = 0.5]{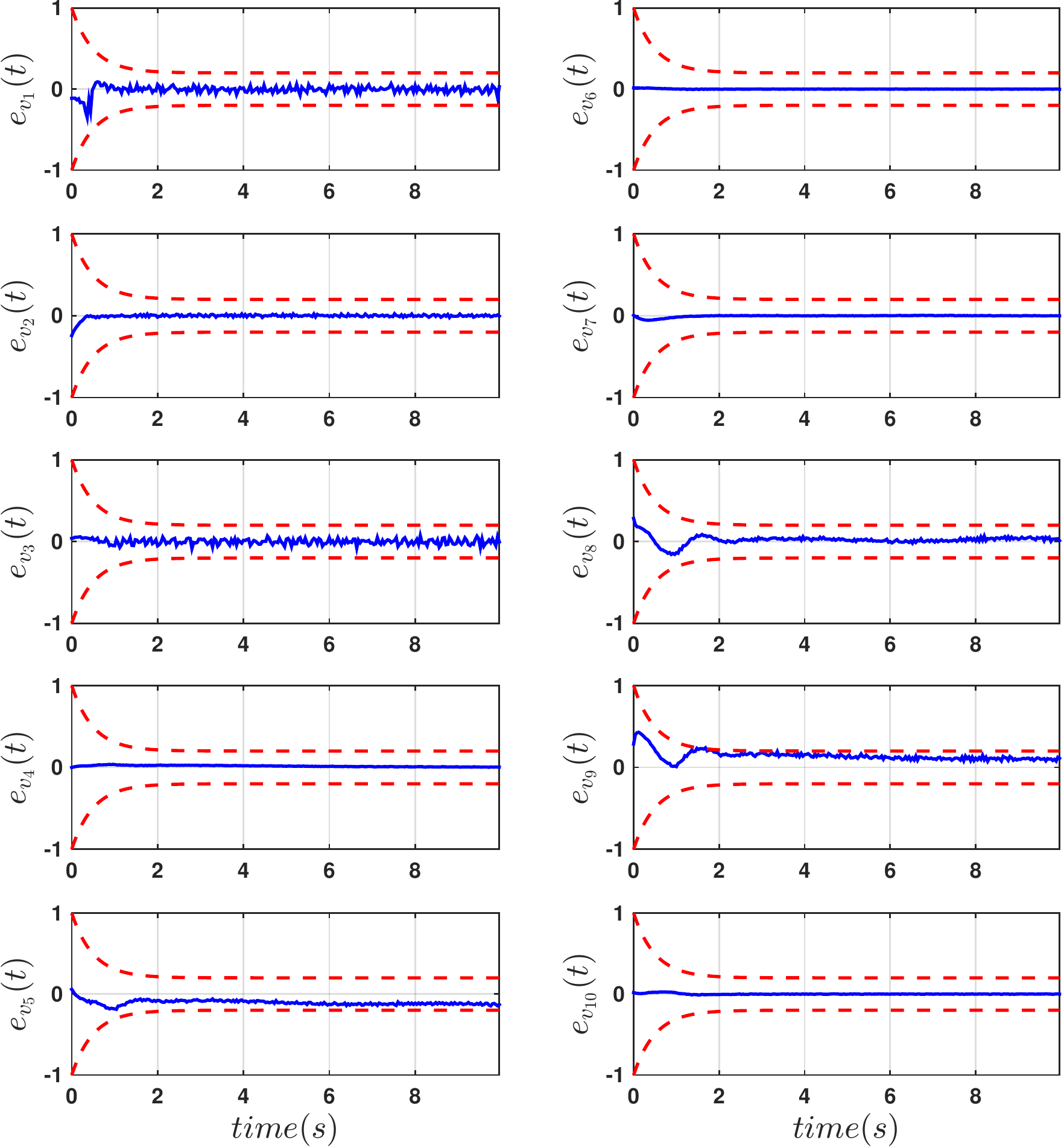}
	\caption{Trajectory scenario: The evolution of the errors at the second level of the proposed control scheme. The errors and performance bounds are indicated by blue and red color respectively.}
	\label{fig:ppv_force}       
\end{figure}
\section{Conclusions and Future Work}
This work presents a robust force/position control scheme for a UVMS in interaction with a compliant environment, which could have direct applications in the underwater robotics (e.g. sampling of the sea organisms, underwater welding, pushing a button).  Our proposed control scheme does not required any priori knowledge of the UVMS dynamical parameters as well as environment model. It guarantees a predefined behavior of the system in terms of desired overshot and transient and steady state performance. Moreover, the proposed control scheme is robust with respect to the external disturbances and measurement noises.  The proposed controller of this work exhibits the following important characteristics: i) it is of low complexity and thus it can be used effectively in most of today UVMS. ii) The performance of the proposed scheme (e.g. desired overshot, steady state performance of the systems) is a priori and explicitly imposed by certain designer-specified performance functions, and is fully decoupled by the control gains selection, thus simplifying the control design. The simulations results demonstrated the efficiency of the proposed control scheme. Finally, future research efforts will be devoted towards addressing the torque controller as well as conducting experiments with a real UVMS system.

\bibliography{mybibfilealina}
\bibliographystyle{ieeetr}

\end{document}